\documentclass{article}

\usepackage{microtype}
\usepackage{graphicx}
\usepackage{subcaption}
\usepackage{booktabs} 

\usepackage{hyperref}

\usepackage[accepted]{MOSS_camera_ready}

\usepackage{amsmath}
\usepackage{amssymb}
\usepackage{mathtools}
\usepackage{amsthm}

\usepackage[capitalize,noabbrev]{cleveref}

\graphicspath{{.}}
\theoremstyle{plain}

\theoremstyle{definition}

\theoremstyle{remark}

\usepackage[textsize=tiny]{todonotes}

\icmltitlerunning{SynDaCaTE: A Synthetic Dataset For Evaluating Part-Whole Hierarchical Inference}

\begin{document}

\onecolumn
\icmltitle{\texorpdfstring{SynDaCaTE: A Synthetic Dataset For Evaluating \\ Part-Whole Hierarchical Inference}{SynDaCaTE: A Synthetic Dataset For Evaluating Part-Whole Hierarchical Inference}}



\icmlsetsymbol{equal}{*}

\begin{icmlauthorlist}
\icmlauthor{Jake Levi}{csox}
\icmlauthor{Mark van der Wilk}{csox}
\end{icmlauthorlist}

\icmlaffiliation{csox}{Department of Computer Science, University of Oxford}

\icmlcorrespondingauthor{Jake Levi}{jake.levi@stcatz.ox.ac.uk}

\icmlkeywords{Synthetic dataset, part-whole hierarchy, inductive bias, data efficiency, capsule models}


\vskip 0.3in


\printAffiliationsAndNotice{}  

\begin{abstract}
Learning to infer object representations, and in particular part-whole hierarchies, has been the focus of extensive research in computer vision, in pursuit of improving data efficiency, systematic generalisation, and robustness. Models which are \emph{designed} to infer part-whole hierarchies, often referred to as capsule networks, are typically trained end-to-end on supervised tasks such as object classification, in which case it is difficult to evaluate whether such a model \emph{actually} learns to infer part-whole hierarchies, as claimed. To address this difficulty, we present a SYNthetic DAtaset for CApsule Testing and Evaluation, abbreviated as SynDaCaTE, and establish its utility by (1) demonstrating the precise bottleneck in a prominent existing capsule model, and (2) demonstrating that permutation-equivariant self-attention is highly effective for parts-to-wholes inference, which motivates future directions for designing effective inductive biases for computer vision.
\end{abstract}

\section{Introduction}
\label{section:Introduction}
In recent years, a dominant trend in machine learning research has been towards finding model architectures that improve with scale \cite{vaswani2017attention}, and scaling them up \cite{sutton2019bitter} with billions of parameters \cite{brown2020language}, trillions of tokens \cite{yang2024qwen2}, hundreds of millions of dollars \cite{maslej2025artificialintelligenceindexreport}, and significant environmental cost \cite{bengio2025international}. There are now strong and increasingly growing financial and environmental incentives to develop machine learning algorithms that achieve similar performance more efficiently.

It is well established that data efficiency can be improved by using more appropriate \textbf{inductive biases} \cite{du2018many, li2021why}, and that models with better data efficiency typically also train faster \cite{hardt2016train}. Developing better inductive biases is therefore a promising strategy for improving the general efficiency of machine learning algorithms. Substantial evidence from the cognitive sciences suggests that object representations \cite{kahneman1992reviewing} and in particular part-whole hierarchies \cite{biederman1987recognition} are ubiquitous in the human visual system. This has further motivated extensive research in computer vision focusing on object-centric learning \cite{burgess2019monet,locatello2020object} and models which are designed to learn part-whole hierarchies \cite{sabour2017dynamic,hinton2018matrix}, commonly referred to as capsule networks \cite{hinton2011transforming}.

Despite much early excitement and published work focusing on capsule networks, they have largely been forgotten in favour of convolutional \cite{he2016deep,liu2022convnet} and attention-based \cite{dosovitskiy2020image,liu2021swin} vision models. Does this mean that exploring visual part-whole hierarchies as an inductive bias for improving data efficiency is a hopeless endeavour? There is one reason to be hopeful. While it is \emph{claimed} that capsule models learn to infer part-whole hierarchies, to the best of our knowledge, there is no empirical evidence to suggest that they \emph{actually} do this efficiently. Before we can determine whether part-whole hierarchies are a useful inductive bias for improving data efficiency in computer vision, we must first (1) define precisely \emph{what it means} to infer a part-whole hierarchy, and (2) establish a procedure to determine which models have the \emph{capacity} to efficiently learn to infer part-whole hierarchies.

In order to evaluate the accuracy with which a model can infer part-whole hierarchies, we must have access to ground-truth part information, which is typically not available in natural visual datasets. We could manually label an existing visual dataset with part information, but this would be undesirably expensive and ambiguous. Instead, we introduce a synthetic dataset that has optional ground-truth part-information built in. Our proposed dataset is \emph{elementary}, but nonetheless it enables us to draw \emph{unique conclusions}, such as pinpointing the bottleneck in an existing capsule model, and motivating future directions for designing effective inductive biases. More specifically, we make the following \textbf{primary contributions}:

\begin{enumerate}
    \item (\S\ref{section:Methods/A Framework For Mereological Inference}) We introduce a framework which clarifies the meaning of part-whole hierarchical inference.
    \item (\S\ref{section:Methods/The SynDaCaTE Dataset}) We present a SYNthetic DAtaset for CApsule Testing and Evaluation, abbreviated as SynDaCaTE.
    \item (\S\ref{section:Results}) We use our SynDaCaTE dataset to demonstrate that:
    \begin{enumerate}
        \item The bottleneck in an existing capsule model \cite{sabour2017dynamic} is inferring parts from an image (rather than inferring wholes from parts), which has implications for designing capsule networks (discussed further in \S\ref{section:Discussion}).
        \item Given explicit part-information, this capsule model is no more efficient than a CNN at inferring wholes from parts.
        \item A permutation-equivariant SetTransformer \cite{lee2019set} is a strong baseline for inferring wholes from parts, which motivates future directions for designing effective inductive biases (discussed further in \S\ref{appendix:Future work}).
    \end{enumerate}
\end{enumerate}


\section{Methods}
\label{section:Methods}
\subsection{A Framework For Mereological Inference}
\label{section:Methods/A Framework For Mereological Inference}

In general, an image may depict a set of several top-level objects, each of which may be considered a ``whole'' object containing a set of ``parts''. Each of those parts may be considered a whole containing its own set of parts, and so on for several layers in a ``part-whole hierarchy''. Each object (part or whole) in an image may be described by a discrete ``class'' label and a continuous ``pose'' vector. The class describes the type of the object, and the pose describes everything which is needed to render the object as seen in the image given its class, which might include $(x,y)$-position, size, rotation, brightness, and so on. In this simplified framework we do not consider noise, or interactions between objects such as reflection, illumination, or shadow. We define the \textbf{generalised pose} of an object to be a one-hot encoding of its class concatenated with its pose. We will say ``inferring an object'' as a shorthand for inferring its generalised pose\footnote{Inferring generalised pose is more general than object classification, detection, and segmentation. A class, bounding box, and segmentation mask could all be computed from a generalised pose, however the generalised pose (in general) contains more information than is available in those labels.}. We define \textbf{inferring a part-whole hierarchy} as inferring all objects (wholes and parts) from an image.

The formal study of part-whole hierarchies is known as \emph{mereology} \cite{cotnoir2021mereology}, and we will refer to inferring part-whole hierarchies as \emph{mereological inference}. We will refer to models which have the capacity to efficiently learn mereological inference as having \emph{mereological capacity}. In general, a model which has mereological capacity \emph{must} be able to efficiently learn to solve two distinct sub-tasks:

\begin{enumerate}
    \item \textbf{Image-to-parts}: infer a set of parts $\mathcal{P}$ from an image $I\in\mathbb{R}^{C \times H \times W}$.
    \item \textbf{Parts-to-wholes}: infer a set of wholes $\mathcal{W}$ from a set of parts $\mathcal{P}$.
\end{enumerate}

\subsection{The SynDaCaTE Dataset}
\label{section:Methods/The SynDaCaTE Dataset}

As previously mentioned, we want to determine which models have mereological capacity, and in order to do so we need ground-truth part information. To this end, we introduce a SYNthetic DAtaset for CApsule Testing and Evaluation, abbreviated as SynDaCaTE. Example images from different SynDaCaTE tasks are shown in Figure \ref{fig:syndacate} in \S\ref{appendix:SynDaCaTE Specification}.

In total across all SynDaCaTE tasks, there are 21 types of object in three categories, which are lines, characters, and words. A full description of the dimensionality and meaning of the poses of objects in each category is presented in \S\ref{appendix:SynDaCaTE Specification/Object Types}.

The sampling procedure for images in SynDaCaTE is described in \S\ref{appendix:SynDaCaTE Specification/Sampling Procedure}. When each image is sampled, we can control various parameters (such as the maximum and minimum number of objects in an image), and observe various types of data about objects, parts, and images. By appropriately controlling parameters and computing inputs and targets from the available data, we can define many different tasks, which can be used to explore a variety of different research questions. We consider several descriptively named tasks including ImToClass, ImToParts, PreTrainedPartsToClass, PartsToChars, and PartsToClass. All tasks have 60k synthetically generated training samples and 10k test samples. A full description of these and some other tasks, including the type and meaning of inputs and targets in all tasks, is included in \S\ref{appendix:SynDaCaTE Specification/Tasks}.

\subsection{Models}
\label{section:Methods/Models}

Vision models (including capsule networks) are most commonly evaluated on object classification, so we start in \S\ref{section:Results} by considering ImToClass. For ImToClass and PreTrainedPartsToClass we will compare a prominent capsule architecture \cite{sabour2017dynamic} against a lightweight modernised CNN architecture, containing a CoordConv layer \cite{liu2018intriguing} followed by a CNN with ReZero residual connections \cite{he2016deep, bachlechner2021rezero}, depthwise-separable convolutions \cite{chollet2017xception} with zero-padding to preserve dimensions, inverted bottlenecks \cite{vaswani2017attention, liu2022convnet}, and a stage-design \cite{liu2022convnet} which divides the network into stages, each of which starts with a strided dense convolution followed by several residual blocks. The final stage is followed by global average pooling and a linear layer.

For the ImToParts task we train the same modernised CNN architecture, except without global average pooling.

We study full parts-to-wholes inference in isolation using the PartsToChars task, and compare a SetTransformer \cite{lee2019set}, a modified DeepSet \cite{zaheer2017deep} which predicts a separate output element for each input element (using a permutation-invariant embedding of the input set) that we refer to as DeepSetToSet, an MLP which predicts output set elements independently as an element-wise function of the input set, and an MLP which predicts the flattened output set as a function of the flattened input set. Both MLPs use ReZero residual connections \cite{he2016deep, bachlechner2021rezero}.

We relate our previous results using PartsToClass, comparing a SetTransformer and the flattened MLP. The SetTransformer uses global average pooling across the set-dimension after the final attention and MLP blocks, followed by a linear layer.

\subsection{Loss Functions}
\label{section:Methods/Loss Functions}

For ImToClass, PreTrainedPartsToClass, and PartsToClass we use the cross-entropy loss. For ImToParts we use the Chamfer MSE loss for set-prediction \cite{zhang2019deep}. For PartsToChars we use the MSE loss averaged over the output set.

\section{Experiments}
\label{section:Results}
\begin{figure}
    \begin{subfigure}{0.498\textwidth}
        \includegraphics[width=\textwidth]{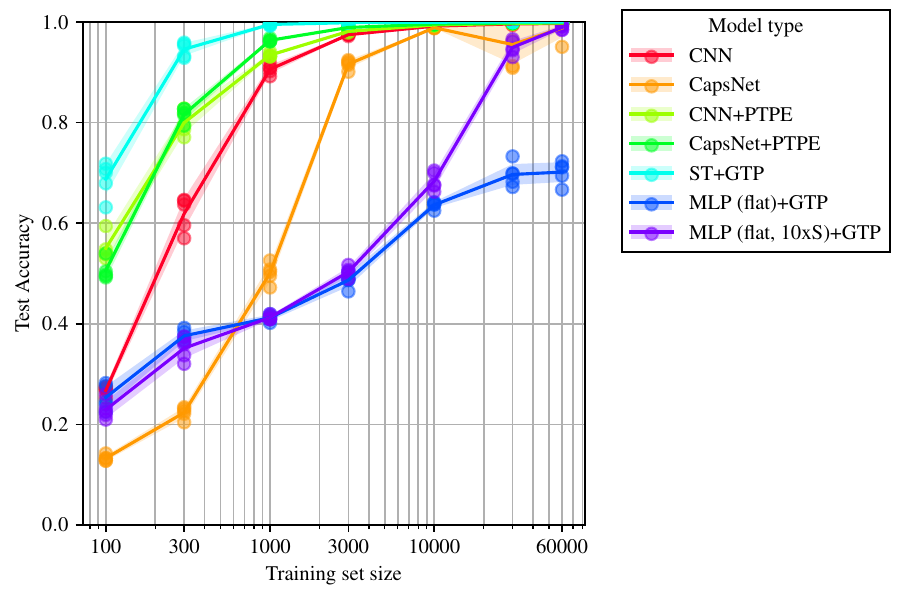}
    \end{subfigure}
    \hfill
    \begin{subfigure}{0.498\textwidth}
        \includegraphics[width=\textwidth]{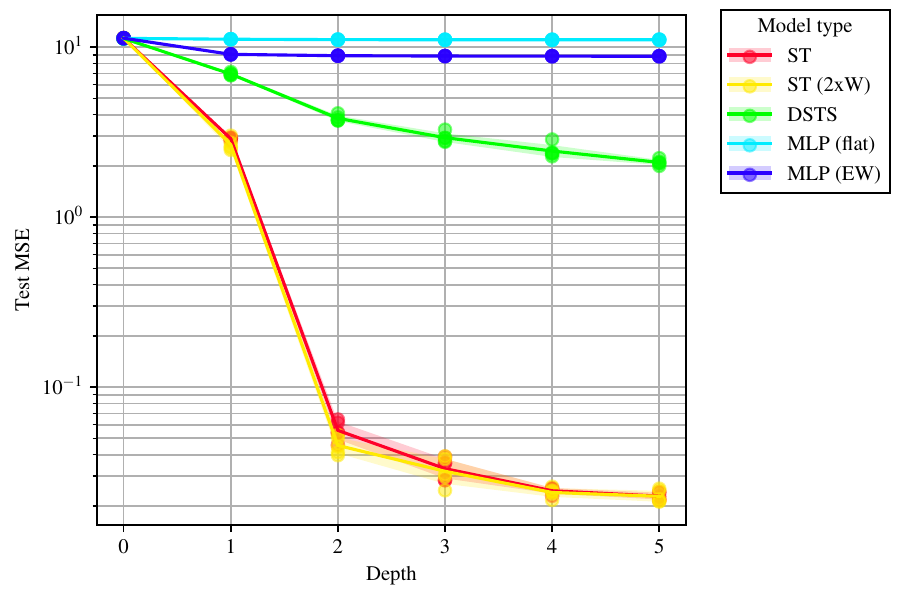}
    \end{subfigure}
    \caption{\textbf{Left}: comparing data efficiency of various models trained on single-object classifcation, from images (ImToClass), pre-trained part encodings (PreTrainedPartsToClass, ``+PTPE''), and ground-truth sets of parts (PartsToClass, ``+GTP''). \textbf{Right}: comparing MSE of various models as a function of depth, trained to predict sets of wholes from sets of parts (PartsToChars). \textbf{Abbreviations}: pre-trained part encodings (PTPE), ground-truth parts (GTP), SetTransformer (ST), DeepSetToSet (DSTS), element-wise (EW), $10\times$ gradient steps (10xS), $2\times$ width (2xW). All experiments are repeated with 5 different random seeds.}
    \label{fig:classification data efficiency,parts-to-wholes MSE vs depth}
\end{figure}

\paragraph{ImToClass.} We start by comparing the data-efficiency of our modernised CNN against CapsNet \cite{sabour2017dynamic} in single-object classification. Both models are trained on the ImToClass task for 5k gradient steps, on subsets of the training set with sample sizes ranging from 100 to 60k. The results are shown in red and orange in Figure \ref{fig:classification data efficiency,parts-to-wholes MSE vs depth} (left). Immediately we can see that our CNN is more data-efficient than CapsNet, achieving significantly higher accuracy at smaller dataset sizes.

\paragraph{ImToParts.} Given that the CapsNet fails to learn efficiently from small sample sizes, we want to understand if the CapsNet is failing in image-to-parts inference, parts-to-wholes inference, or both. To this end, we train CNNs to predict sets of part poses from an image. We train each CNN for 100 epochs on the ImToParts task, using the full 60k training set. Across 5 random seeds, the CNNs reach an average test-set MSE of $0.0269\pm 0.0026$, with the best model reaching 0.02329. We can qualitatively assess the performance of a model trained on ImToParts by evaluating the model on an image, feeding the predictions of the model to the rendering pipeline used to create the dataset, and comparing against the ground-truth targets. The predictions of the best CNN trained on ImToParts are shown in Figure \ref{fig:CNN ImToParts predictions} in \S\ref{appendix:Additional results}, and are reasonably accurate, although not perfect. We create the PreTrainedPartsToClass task by iterating through ImToClass, feeding each image through the best CNN trained on ImToParts, and replacing the image with the activations in the final hidden layer of the CNN.

\paragraph{PreTrainedPartsToClass.} Figure \ref{fig:classification data efficiency,parts-to-wholes MSE vs depth} (left) also shows the data-efficiency of (randomly initialised) CNNs and CapsNets trained on PreTrainedPartsToClass for 5k gradient steps, in light green and dark green. When given access to part-information, both models are significantly more data-efficient compared to training from images, and also have very similar performance to each other, which suggests at least three important conclusions:

\begin{enumerate}
    \item The bottleneck in CapsNet is inferring \emph{parts from images}, rather than inferring wholes from parts (because the decrease in performance after removing pre-trained \emph{part-information} is much worse in CapsNet than in CNN).
    \item A naive capsule model (CapsNet) is not substantially more efficient at learning parts-to-wholes inference than a CNN.
    \item Part-information provides a useful representation for data-efficient classification, even though the part-representations are noisy, and contain no explicit class-information.
\end{enumerate}

\paragraph{PartsToChars.} Should we expect that, in general, a CNN (or a CapsNet which is not substantially better) has \emph{the best} possible inductive bias for inferring wholes from parts? We begin to explore this question by turning to the PartsToChars task, in which the input is a randomly ordered ground-truth set of poses of parts of between one and three objects, and the target is the set of generalised poses of the objects. The input is designed to represent some of the challenges faced by the receptive fields of neurons. Firstly, as an object transforms relative to a retina or camera (for example by rotating or moving closer), part-representations may move into the receptive fields of other neurons in more complex ways than simple translation. Secondly, the receptive field of any neuron might observe parts belonging to multiple distinct objects.

Because the input is a randomly ordered set, we primarily compare set-function models trained on this task. We want to know which inductive bias is most effective for this task, and for any model, it is useful to know how performance varies with important hyperparameters such as depth. To this end, we plot test MSE as a function of depth for several models in Figure \ref{fig:classification data efficiency,parts-to-wholes MSE vs depth} (right). All models are trained for 100 epochs. The SetTransformer is strikingly effective at this task, outperforming all other baselines by more than an order of magnitude at depths $\ge2$, but \emph{not} with a single layer of self-attention. Does the SetTransformer rely on some fundamentally useful computation that \emph{needs} $\ge2$ layers of self-attention, or does the shallow SetTransformer simply not have enough parameters? We address this question by sweeping over a SetTransformer with double the width ($\approx4\times$ parameters), also shown in Figure \ref{fig:classification data efficiency,parts-to-wholes MSE vs depth} (right). Despite the extra parameters, the performance is remarkably insensitive to width, which suggests that $\ge2$ self-attention layers \emph{are} in some way fundamentally useful. Future work may examine this through the lens of mechanistic interpretability, which has shown that ``Induction heads'' also arise in transformers with no fewer than two attention layers \cite{elhage2021mathematical}. Rendered predictions from the best SetTransformer and DeepSetToSet trained on PartsToChars are displayed in Figures \ref{fig:ST PartsToChars predictions} and \ref{fig:DSTS PartsToChars predictions} in \S\ref{appendix:Additional results}.

\paragraph{PartsToClass.} Finally, we return to data-efficiency and single-object classification, by comparing the performance of a SetTransformer and a flattened MLP on PartsToClass. The results for various training sample sizes after training for 5k gradient steps are shown in Figure \ref{fig:classification data efficiency,parts-to-wholes MSE vs depth} (left). Once again, the SetTransformer outperforms all other baselines, especially for small training sample sizes. The flattened MLP with 5k gradient steps does not achieve good accuracy for any training sample size. Using 50k gradient steps improves performance for larger but not smaller training sample sizes, which aligns with theoretical results connecting generalisation performance to convergence speed \cite{hardt2016train}.

\section{Conclusion}
\label{section:Discussion}
Although our proposed SynDaCaTE dataset is elementary, its thoughtful design has allowed us to draw unique conclusions about an existing capsule model, and about the effectiveness of permutation-equivariant SetTransformers in parts-to-wholes inference. Our conclusions about CapsNet have strong implications for future work exploring capsule networks, including that image-to-parts inference should be considered explicitly and empirically (possibly using our dataset) and not taken for granted. Our findings about parts-to-wholes inference with SetTransformers motivate future directions for designing effective inductive biases for computer vision, which (because of space constraints) we discuss further in \S\ref{appendix:Future work}.

\section*{Reproducability}

Our code is available at $<$\url{https://github.com/jakelevi1996/syndacate-public}$>$. We provide a full description of hyperparameters in \S\ref{appendix:Hyperparameters}.

\section*{Acknowledgements}

We would like to thank Christian Rupprecht and Christian Schroeder de Witt for their helpful feedback, discussion, and advice. We would like to thank the reviewers for their helpful reviews. This work was supported by the EPSRC Centre for Doctoral Training in Autonomous Intelligent Machines and Systems (grant number EP/S024050/1).

\bibliography{references}
\bibliographystyle{icml2025}

\appendix
\newpage
\section{Future work}
\label{appendix:Future work}

\paragraph{Overview.} Our findings in \S\ref{section:Results} (which were facilitated by our proposed synthetic dataset) motivate several promising directions for future work. We first discuss \emph{intuition} for designing effective inductive biases for computer vision, followed by possible extensions to the SynDaCaTE dataset.

\paragraph{Visual Inductive Biases.} We demonstrated in \S\ref{section:Results} that a permutation-invariant SetTransformer was significantly more data-efficient than other baselines at classifying objects from a ground-truth \emph{set} of parts (which was designed to represent the receptive field of a visual neuron). However, a general-purpose vision model must take its input from images. How should we best encompass our findings in a general-purpose vision model? A simple approach would be to use our lightweight modernised CNN as a starting point, and replace depthwise-separable convolutions with sliding-window self-attention\footnote{In sliding-window self-attention, each pixel is used as a query for keys and values in a local neighbourhood of pixels.} and point-wise MLPs in alternating residual blocks. These changes maintain the overall structure of a CNN, while modifying the inductive bias only within local neighbourhoods. The resulting model resembles Stand-Alone Self-Attention \cite{ramachandran2019stand} without normalisation layers \cite{ioffe2015batch} or layer-wise relative positional embeddings \cite{shaw2018self} and with ReZero residual connections \cite{ramesh2021zero} and an initial CoordConv layer \cite{liu2018intriguing} instead. Strided dense convolutions for inter-stage downsampling could be replaced with local linear average pooling, or local attention-weighted average pooling. Our preliminary experiments with this model, which we refer to as MereoFormer, demonstrate promising mereological capacity, however a remaining practical challenge is to improve the efficiency of our software implementation.

\paragraph{Extending SynDaCaTE.} The \emph{simplicity} of our SynDaCaTE dataset has allowed us to draw fundamental scientific conclusions without superfluous additional complexity. In future, we will want to draw analogous conclusions about more complex visual environments which are more relevant to practical real-world applications of computer vision. This motivates developing the SynDaCaTE dataset in several directions:

\begin{enumerate}
    \item \textbf{Generalisation}: The training and test splits are currently sampled from the same distribution. These distributions could be structurally modified to support investigations into systematic \cite{bahdanau2018systematic} and compositional \cite{wiedemer2023provable} generalisation.
    \item \textbf{Classes}: The SynDaCaTE dataset contains a small number of object classes. More object classes (both wholes and parts) could be added, which may also support investigating meta-learning and transfer-learning of new object classes.
    \item \textbf{3D}: The SynDaCaTE dataset is purely 2D, whereas the natural visual world is 3D. Synthetic 3D part-whole hierarchies could be designed and implemented using a 3D rendering pipeline, to explore the extent to which mereological capacity of various models is consistent between 2D and 3D visual environments.
    \item \textbf{Video}: SynDaCaTE contains only static images, and our experiments only considered models which were trained on supervised learning tasks. The SynDaCaTE dataset could be extended to include \emph{videos} of moving objects, which may be used to explore the extent to which models which have mereological capacity are also able to learn meaningful representations of parts and wholes (which for example could be linearly decoded from hidden representations) from \emph{unsupervised} next-frame video prediction. We might refer to such a dataset as ``Video SynDaCaTE'', or more concisely as ``VinDaCaTE''.
    \item \textbf{Interaction}: VinDaCaTE could be designed to optionally include interactions between objects in a video, which could be used to systematically investigate how well various models are able to learn to understand such interactions.
    \item \textbf{Noise}: All inputs and targets in SynDaCaTE are noise-free by design. Future work may explore the effects of various noise distributions on the data-efficiency and final performance of vision models trained on image-to-parts, parts-to-wholes, and end-to-end learning tasks using the SynDaCaTE dataset.
\end{enumerate}

\newpage
\section{Additional results}
\label{appendix:Additional results}

\begin{figure}[H]
    \centering

    \includegraphics[width=0.9\textwidth]{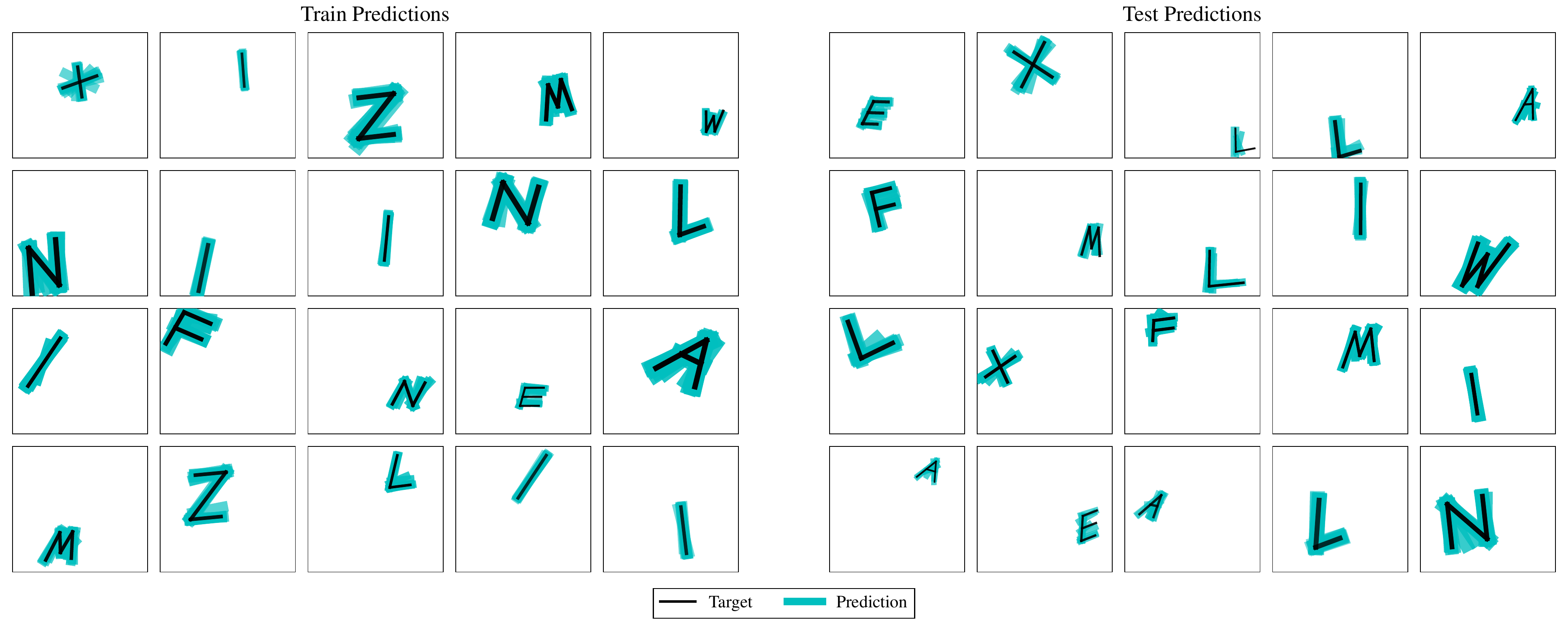}
    \caption{Rendered part-predictions of the best CNN trained on ImToParts.}
    \label{fig:CNN ImToParts predictions}

    \vspace{\floatsep}

    \includegraphics[width=0.9\textwidth]{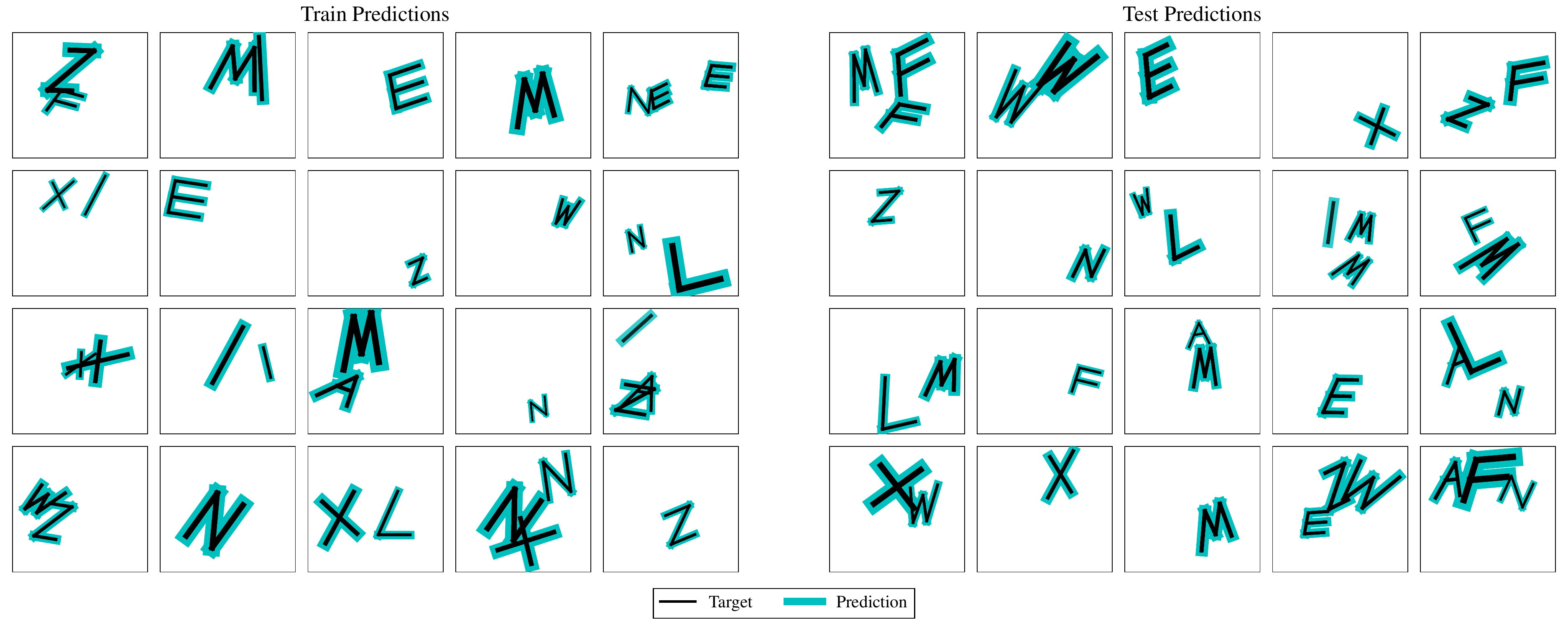}
    \caption{Rendered character-predictions of the best SetTransformer trained on PartsToChars.}
    \label{fig:ST PartsToChars predictions}

    \vspace{\floatsep}

    \includegraphics[width=0.9\textwidth]{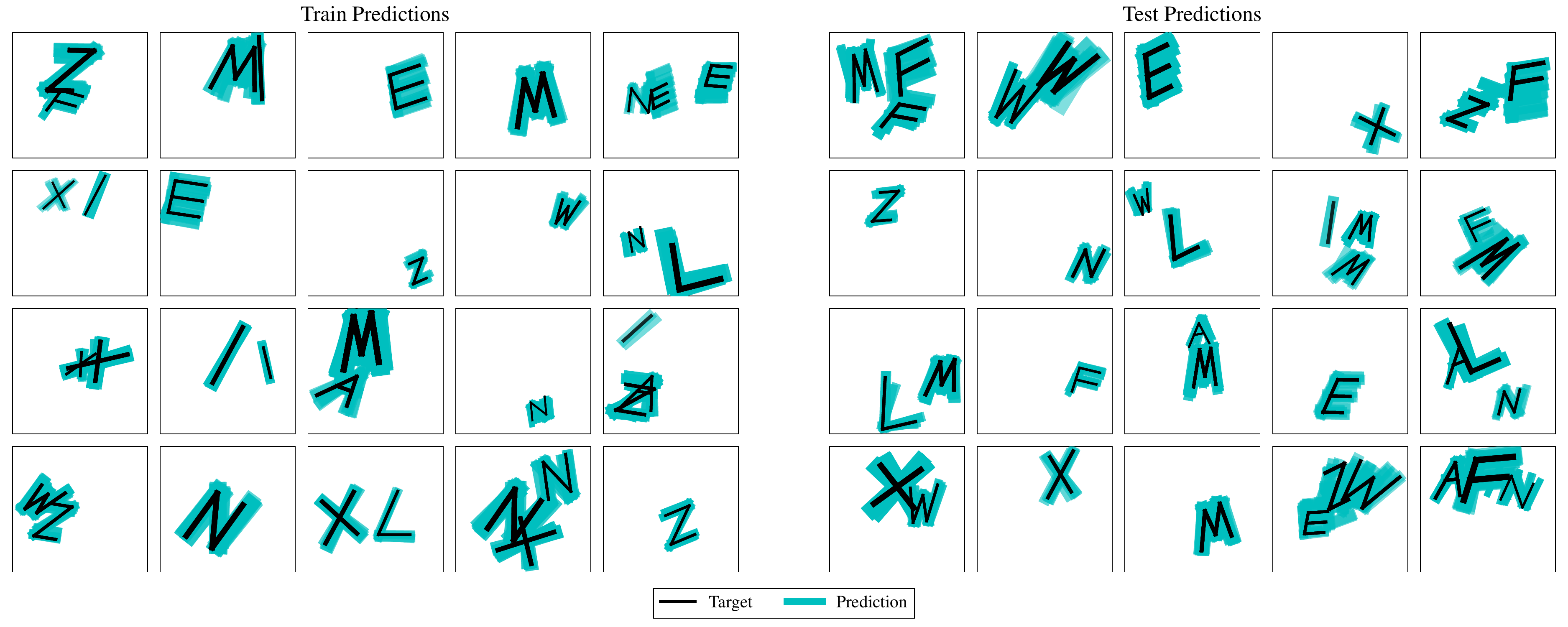}
    \caption{Rendered character-predictions of the best DeepSetToSet trained on PartsToChars.}
    \label{fig:DSTS PartsToChars predictions}
\end{figure}

\newpage
\section{SynDaCaTE Specification}
\label{appendix:SynDaCaTE Specification}

\begin{figure}
    \centering
    \includegraphics[width=\textwidth]{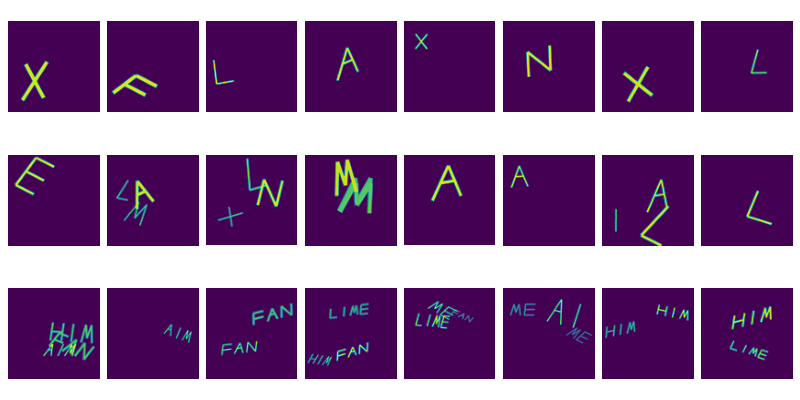}
    \caption{Example images from three different SynDaCaTE tasks. \textbf{Top row}: ImToClass. \textbf{Middle row}: ImToChars. \textbf{Bottom row}: Words.}
    \label{fig:syndacate}
\end{figure}

\subsection{Object Types}
\label{appendix:SynDaCaTE Specification/Object Types}

Across all tasks in SynDaCaTE, there are 21 types of object in three categories:

\begin{enumerate}
    \item \textbf{Lines}: these are the bottom-level parts in all tasks. There is only one type of line. Each line has a 6D pose-vector, with dimensions referring to $x$ and $y$ coordinates of both endpoints, thickness, and brightness. When the ground-truth set of parts is available, each line is included twice, once for each permutation of its endpoints. This is because both permutations of endpoints render identical images, so a single permutation would not be identifiable from a rendered image.
    \item \textbf{Characters}: these are composite objects composed of lines. There are 10 types of characters. Each character has an 8D pose-vector, with dimensions referring to $x$-position, $y$-position, scale, rotation, wideness, italic, line-thickness, and brightness.
    \item \textbf{Words}: these are composite objects composed of characters. There are 10 types of words. The dimensionality and meaning of the pose vector of a word is the same as for a character.
\end{enumerate}

\subsection{Sampling Procedure}
\label{appendix:SynDaCaTE Specification/Sampling Procedure}

Each image is sampled according to the following procedure:

\begin{enumerate}
    \item Sample the number of top-level objects in the image.
    \item For each object:
    \begin{enumerate}
        \item Sample a discrete class label $c_i$ and continuous pose vector $p_i$.
        \item Initialise a set of parts (in some canonical pose) as a function of $c_i$.
        \item Transform each part as a function of $p_i$.
        \item Recursively initialise the sub-parts of each part until reaching bottom-level parts (lines).
    \end{enumerate}
    \item Aggregate all bottom-level parts from all top-level objects into a single sequence and sort it by depth (in SynDaCaTE we simply equate depth with inverse brightness, so brighter parts are in front).
    \item Render the sorted parts in order to form an image.
\end{enumerate}

\subsection{Tasks}
\label{appendix:SynDaCaTE Specification/Tasks}

We present 7 supervised learning tasks, with inputs $x$ and targets $t$ as follows:

\begin{enumerate}
    \item \textbf{ImToParts}: $x \in [0, 1]^{1 \times 100 \times 100}$ is an image of a single character, $t \in \mathbb{R}^{9 \times 6}$ is the set of poses of the parts (lines) in the character. The maximum number of parts per character is 4, each part is represented twice (see above), and the total number of part-poses is padded with zeros up to 9 to ensure that $t$ always contains at least one all-zero pose vector. $t$ contains only 6D poses and no class information (because there is only one type of line). The order of parts in $t$ is randomised.
    \item \textbf{PartsToChars}: $x \in \mathbb{R}^{25 \times 6}$ is a set of poses of the parts of between 1 and 3 characters, $t \in \mathbb{R}^{25 \times 18}$ contains the generalised pose of the character that each part belongs to. The outer dimensions of $x$ and $t$ are padded with zeros up to 25. The order of parts in $x$ is randomised, and the order of characters in $t$ is aligned to the order in $x$. Each generalised pose in $t$ is repeated according to its number of parts.
    \item \textbf{ImToChars}: $x \in [0, 1]^{1 \times 100 \times 100}$ is an image of between 1 and 3 characters, $t \in \mathbb{R}^{4 \times 18}$ contains the generalised poses of the characters, padded with zeros up to an outer dimension of 4. The order of parts in $t$ is randomised.
    \item \textbf{ImToClass}: $x \in [0, 1]^{1 \times 100 \times 100}$ is an image of a single character, $t \in \{1, \dots, 10\}$ is the class of the character (represented as an integer). Pose information is discarded.
    \item \textbf{PartsToClass}: $x \in \mathbb{R}^{9 \times 6}$ is the set of poses of the parts of a single character, $t \in \{1, \dots, 10\}$ is the class of the character.
    \item \textbf{PreTrainedPartsToClass}: $x \in [0, 1]^{M \times H^\prime \times W^\prime}$ is the activations in the final hidden layer of a model trained on ImToParts, frozen, and evaluated on an image of a single character, $t \in \{1, \dots, 10\}$ is the class of the character.
    \item \textbf{Words}: $x \in [0, 1]^{1 \times 100 \times 100}$ is an image of between 1 and 3 words, $t \in \mathbb{R}^{4 \times 18}$ contains the generalised poses of the words, padded with zeros up to an outer dimension of 4.
\end{enumerate}

\newpage
\section{Hyperparameters}
\label{appendix:Hyperparameters}

\subsection{All Models}

\begin{center}
    \begin{tabular}{|c|c|}
        \hline
        Optimiser & Adam \cite{kingma2014adam} \\
        Initial learning rate & 1e-3 \\
        Final learning rate & 1e-5 \\
        Learning rate schedule & Cosine decay \\
        Batch size & 100 \\
        \hline
    \end{tabular}
\end{center}

\subsection{Data-efficiency Classification Sweeps}

\begin{center}
    \begin{tabular}{|c|c|c|}
        \hline
        CNN (ImToClass) & Kernel size & 5 \\
        & Model dimension & 64 \\
        & Expand ratio (in residual layers) & 2 \\
        & Number of stages & 3 \\
        & Number of blocks per stage & 2 \\
        & Stride per stage & 2 \\
        & Input embedding & CoordConv \cite{liu2018intriguing} \\
        \hline
        CapsNet (ImToClass) & Routing iterations & 3 \\
        & All other hyperparameters & Default* \\
        \hline
        CNN (PreTrainedPartsToClass) & Kernel size & 5 \\
        & Model dimension & 64 \\
        & Expand ratio (in residual layers) & 2 \\
        & Number of stages & 1 \\
        & Number of blocks per stage & 3 \\
        & Stride per stage & 1 \\
        & Input embedding & CoordConv \cite{liu2018intriguing} \\
        \hline
        CapsNet (PreTrainedPartsToClass) & Routing iterations & 3 \\
        & All other hyperparameters & Default* \\
        \hline
        SetTransformer (PartsToClass) & Depth & 5 \\
        & Model dimension & 64 \\
        & Number of heads & 8 \\
        & Expand ratio (in residual layers) & 2 \\
        \hline
        MLP (PartsToClass) & Depth & 5 \\
        & Model dimension & 100 \\
        & Expand ratio (in residual layers) & 2 \\
        \hline
    \end{tabular}
\end{center}

*For CapsNets, we used the public implementation available at \newline$<$\url{https://github.com/adambielski/CapsNet-pytorch}$>$, including default hyperparameters.

\subsection{PartsToChars Depth Sweeps}

\begin{center}
    \begin{tabular}{|c|c|c|}
        \hline
        SetTransformer & Model dimension & 64 \\
        & Number of heads & 8 \\
        & Expand ratio (in residual layers) & 2 \\
        \hline
        SetTransformer (2xW) & Model dimension & 128 \\
        & Number of heads & 8 \\
        & Expand ratio (in residual layers) & 2 \\
        \hline
        DeepSetToSet & Model dimension & 100 \\
        \hline
        MLP & Model dimension & 100 \\
        & Expand ratio (in residual layers) & 2 \\
        \hline
    \end{tabular}
\end{center}

\subsection{ImToParts}

\begin{center}
    \begin{tabular}{|c|c|c|}
        \hline
        CNN & Kernel size & 5 \\
        & Model dimension & 64 \\
        & Expand ratio (in residual layers) & 2 \\
        & Number of stages & 3 \\
        & Number of blocks per stage & 2 \\
        & Stride per stage & 2 \\
        & Input embedding & CoordConv \cite{liu2018intriguing} \\
        \hline
    \end{tabular}
\end{center}

\end{document}